\documentclass[conference]{IEEEtran}

\usepackage{fancyhdr}
\IEEEoverridecommandlockouts
\usepackage{graphicx} 
\usepackage{pifont}
\usepackage{amsmath}
\usepackage{amssymb}  
\usepackage{amsfonts}
\usepackage{algorithmic}
\usepackage{array}
\usepackage{multicol}

\usepackage{textcomp}
\usepackage{stfloats}
\usepackage{url}
\usepackage{verbatim}
\usepackage{graphicx}
\usepackage{caption}
\usepackage{subcaption}
\usepackage[ruled, vlined, linesnumbered]{algorithm2e}
\usepackage{cite}
\usepackage{booktabs}
\usepackage{amssymb,amsfonts,bm}
\usepackage{amsmath,tikz}
\usepackage{color}
\usepackage{mathtools}
\usepackage[hidelinks]{hyperref} 
\usepackage{rotating}
\usepackage{blkarray}
\usepackage{physics}
\usetikzlibrary{arrows}
\usepackage{makecell}

\usepackage{amsthm}
\usepackage{comment}
\usepackage{multirow}
\usepackage{geometry}
\graphicspath{{images/}}

\usepackage{textcomp}
\usepackage{tikz}
\newcommand\copyrighttext{%
    \footnotesize This work has been accepted to IEEE International Conference on Robotics and Automation (ICRA) © 2025 IEEE. Personal use of this material is permitted.
    Permission from IEEE must be obtained for all other uses, including reprinting/redistribution, 
    creating new works, or reuse of any copyrighted components of this work in other media.}
\newcommand\copyrightnotice{%
    \begin{tikzpicture}[remember picture,overlay]
    \node[anchor=south,yshift=10pt] at (current page.south) {\fbox{\parbox{\dimexpr\textwidth-\fboxsep-\fboxrule\relax}{\copyrighttext}}};
    \end{tikzpicture}%
}
\begin{document}


  
\title{\LARGE \bf
Large-Scale UWB Anchor Calibration and One-Shot Localization Using Gaussian Process \\

\author{Shenghai Yuan$\dagger$, 
Boyang Lou$\dagger$, 
Thien-Minh Nguyen, 
Pengyu Yin,
Muqing Cao,\\
Xinghang Xu,
Jianping Li, 
Jie Xu,
Siyu Chen,
Lihua Xie,~\IEEEmembership{Fellow,~IEEE}}

\thanks{This research is supported by the National Research Foundation, Singapore, under its Medium-Sized Center for Advanced Robotics Technology Innovation (CARTIN) and Delta-NTU Corporate Lab. }
\thanks{$\dagger$ Equal contribution. All authors are with the School of Electrical and Electronic Engineering, Nanyang Technological University, 50 Nanyang Avenue, Singapore 639798, 
   { Email:   \{shyuan, thienminh.nguyen, mqcao, xinhang.xu, jianping.li, elhxie\} @ntu.edu.sg} \{pengyu001\} @e.ntu.edu.sg}%
\thanks{Boyang Lo is also with the Beijing University of Posts and Telecommunications, Beijing, China, 100876, 
   { Email:  woshiluobodan@bupt.edu.cn.}}%
}

\maketitle

\begin{abstract}

Ultra-wideband (UWB) is gaining popularity with devices like AirTags for precise home item localization but faces significant challenges when scaled to large environments like seaports. The main challenges are calibration and localization under obstructed conditions, which are common in logistics environments.
Traditional calibration methods, dependent on line-of-sight (LoS), are slow, costly, and unreliable in seaports and warehouses, making large-scale localization a significant pain point in the industry. To overcome these challenges, we propose a one-shot calibration and localization framework based on UWB-LiDAR fusion. Our method uses Gaussian processes to estimate the anchor position from continuous-time LiDAR Inertial Odometry with sampled UWB ranges. This approach ensures accurate and reliable calibration with only one round of sampling in large-scale areas, i.e., 600x450 $m^2$. With LoS issues, UWB-only localization can be problematic, even when anchor positions are known. We demonstrate that by applying a UWB-range filter, the search range for LiDAR loop closure descriptors is significantly reduced, improving both accuracy and speed. 
This concept can be applied to other loop closure detection methods, enabling cost-effective localization in large-scale warehouses and seaports. It significantly improves precision in challenging environments where the UWB-only and LiDAR-Inertial methods fail, as shown in the video \url{https://youtu.be/oY8jQKdM7lU }. We will open-source our datasets and calibration codes for community use. 

\end{abstract}
\copyrightnotice

\section{Introduction}

Ultra-Wideband (UWB) \cite{nguyen2022ntu} is a short-range, energy-efficient radio communication technology mainly used for precise location detection and relative range measurement \cite{yuan2021survey}.
Devices such as Apple AirTags and Android SmartTags are gaining popularity, with UWB technology that enables them to track and locate household items accurately \cite{yuan2021survey}. 

\begin{figure}[t]
\centering
\includegraphics[width=9.9cm]{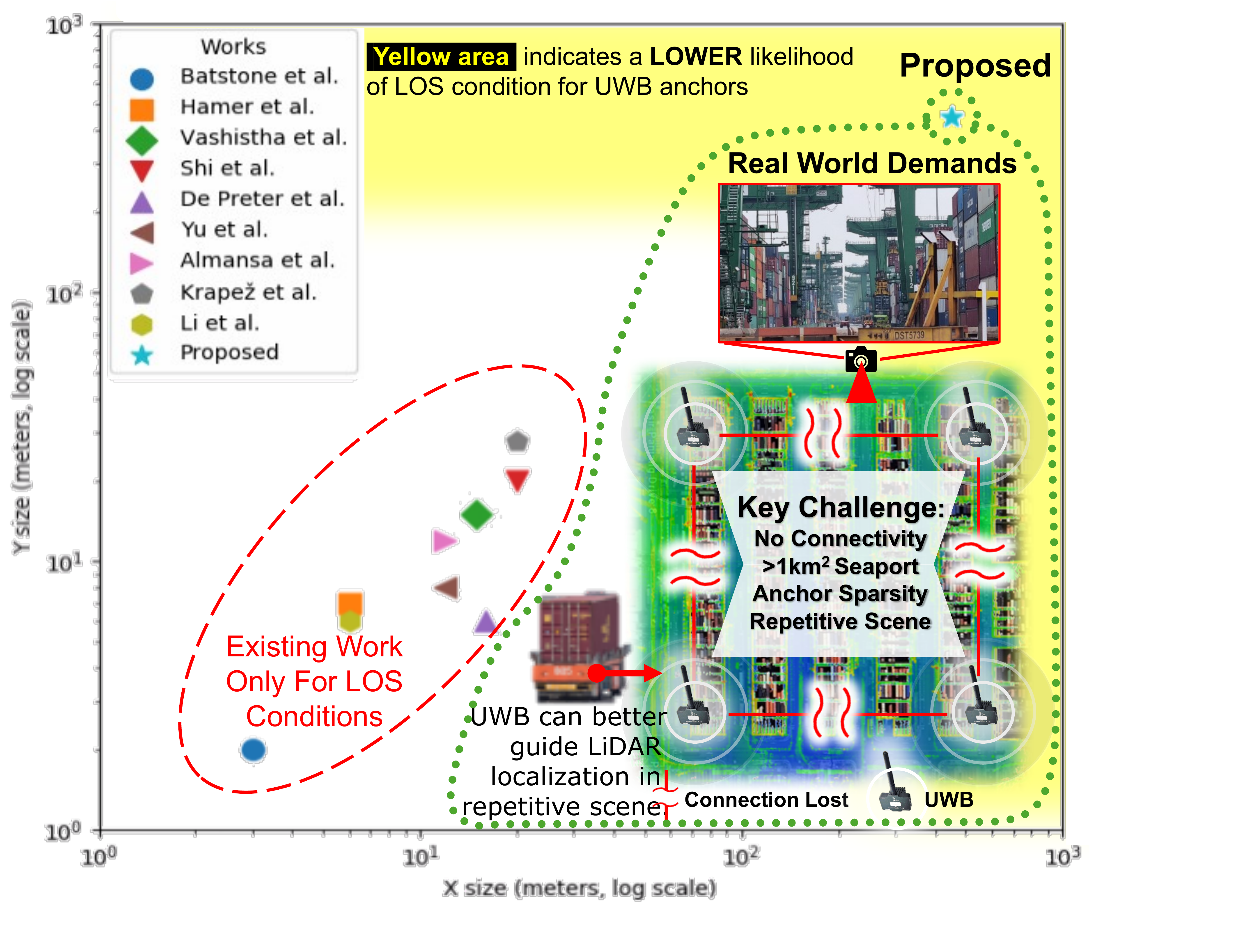}
\vspace{-3.1em}
\caption{\footnotesize Survey of UWB-based localization methods and their maximum coverage size, See Sec. II for detailed explanations.}
\label{fig:comparsion}
\vspace{-2em}
\end{figure}

Although it has seen success in sports-like applications, such as football tracking, UWB still faces significant challenges in large-scale real-time industrial localization applications \cite{bastida2018accuracy}. Most UWB studies are confined to smaller or indoor environments operating in absolute positioning (AP) mode \cite{nguyen2022ntu, vashistha2018self}, as shown in Fig. \ref{fig:comparsion}. In this setup, all anchors must be calibrated \cite{corbalan2023self}, and the tag requires constant range measurements from multiple anchors, which becomes impractical in occluded environments \cite{nguyen2021viral}. This limitation reduces the practicality of UWB in massive areas such as seaports or warehouses, where obstructions present significant hurdles.

Adopting UWB in large outdoor environments \cite{wei2024fusionportablev2,yin2021m2dgr} presents challenges due to frequent Non-Line-of-Sight (NLoS) issues and interference from other devices, which limit the effective sensing range. In the worst case, there could be zero connection between the anchors due to the newly added container, as shown in Fig. \ref{fig:comparsion}. The absence of reliable GPS signals in obstructed environments further complicates anchor calibration and requires alternative approaches \cite{zhan2024fast,hamesse2024fast,nguyen2021liro,yuan2024mmaud,cao2022direct,ji2024lio,xu2024selective} to assist with calibration. Even with calibration \cite{li2025limo,liao2023se,wang2015automatic}, the system struggles to localize \cite{10801455,esfahani2019towards} in NLoS conditions \cite{Nguyen2025ULOC}, creating a significant bottleneck for field robotics and logistics automation.

In this work, we propose a comprehensive approach to address the challenges of large-scale UWB calibration and one-shot localization. Our method leverages Continuous-Time LiDAR-Inertial Odometry (CT-LIO) to generate a trajectory that can be sampled at any point. By interpolating this trajectory, we align each UWB measurement with its corresponding position. The anchor position estimates are iteratively refined using Gaussian Process (GP) \cite{Nguyen2024GPTR} regression, which models spatial dependencies and improves accuracy.  We use GP for calibration instead of traditional Nonlinear Least Squares (NLS), as it allows for flexible modeling of spatially correlated UWB errors and provides a probabilistic framework for uncertainty quantification. Even after calibration, the system remains unable to localize in NLoS mode using standard UWB techniques \cite{li2023continuous}. To overcome this, we combine our solution with a fast descriptor-based method \cite{yuan2023std} to enhance one-shot global localization in large and repetitive environments. Our approach significantly improves existing localization methods and introduces new possibilities previously unexplored.
Our contributions can be summarized as follows:
\begin{enumerate}
  \item   The paper introduces a framework that combines UWB data with CT-LIO and GP to accurately calibrate UWB anchors in large environments using only one round of sampling. This enhances localization in obstructed scenes where UWB or LiDAR alone would fail.
  \item  It presents a one-shot localization process that reduces search time by filtering UWB ranges with existing descriptors, enabling efficient and accurate localization in complex areas without needing multiple UWB anchors, making the solution cost-effective and practical.
  \item The method is validated by real-world experiments in a 600x450 m² environment, showing improved accuracy and reduced processing time, proving its practicality for large-scale applications in challenging conditions.
  \item We will make datasets and calibration codes publicly available for the benefit of the community. \url{https://github.com/luobodan/GP-UWB-calibration}
\end{enumerate}

\vspace{-5pt}

\section{Related Work}
To the best of our knowledge, there are very few works and datasets that address large-scale UWB anchor calibration with NLoS constraint, as it poses a significant technical challenge \cite{corbalan2023self}.

Most of the existing work on UWB only aims to address indoor UWB anchor calibration. Batstone et al. \cite{batstone2017towards} proposed a RANSAC-based self-calibration approach designed to work with minimal solvers to create a robust and efficient solution for the self-calibration problem. However, this approach is tailored for a small-scale scenario, covering only a 3x2m² area. Yu et al.\cite{yu2017self} introduced a theoretical framework based on the Markov State Transition Equation to compute state vectors for all anchor coordinates, along with an iterative trilateral localization technique. Despite its innovation, this approach still heavily relies on Line-of-Sight (LoS) conditions, reflecting an early and somewhat rudimentary approach in UWB self-calibration methods.

Later approaches \cite{hamer2018self} ingeniously incorporate multi-robot message exchange for enhanced UWB self-calibration. These methods leverage network synchronization to integrate the self-calibration process with the localization of mobile nodes. This is achieved by intercepting the self-calibration messages exchanged among the anchors. However, this solution requires additional robots, which can be tedious and costly. Therefore, the newer schemes \cite{vashistha2018self} employ an impulse-radio-based indoor positioning system, utilizing TDOA measurements to autonomously determine anchor positions without requiring prior locations that can work for up to 15m.  Andreas et al.\cite{de2019range} address range bias by modeling and compensating for it, achieving centimeter-level positioning accuracy, and introducing a semi-automated auto-calibration procedure that can work for a 12x8m² area with RTK GPS and good LoS condition.
Shi et al. \cite{shi2019anchor} combine UWB with low-cost IMU sensor data using a SLAM-based self-calibration approach that employs an error-state Kalman filter to fuse UWB and IMU inputs. This method requires only a mobile node to move freely within the area to collect calibration data. However, it is limited to simulations, as high-frequency noise in real-world scenarios causes significant position drift.

Almansa et al. \cite{almansa2020autocalibration} proposed a collaborative auto-calibration algorithm using 4 LoS anchors and least square, with simulated accuracy dropping from 4 cm in a 1x1m² area to 135 cm in a 12x12m² space,  highlighting considerable error even in relatively small spaces. Krapež et al. \cite{krapevz2020anchor} propose calibration modules using four rigid UWB tags as a combined pattern, similar to the VICON motion capture calibration IR stick, but still heavily reliant on LoS conditions. Van Herbruggen et al. \cite{van2023multihop} proposed a multi-stage self-calibration algorithm using UWB measurements and collaborative localization. The algorithm optimizes with gradient descent and refines with a Kalman Filter. However, the algorithm depends on LoS, works only in small areas, and requires data from at least two other anchors, which significantly limits its scalability.  Li et al. \cite{li2023continuous} recently proposed continuous-time UWB-IMU joint calibration and localization using B-splines, though limited to LoS in small rooms. This approach inspires us to use similar concepts for global localization with new modalities.

\begin{figure*}[t]
\centering
\includegraphics[width=\textwidth]{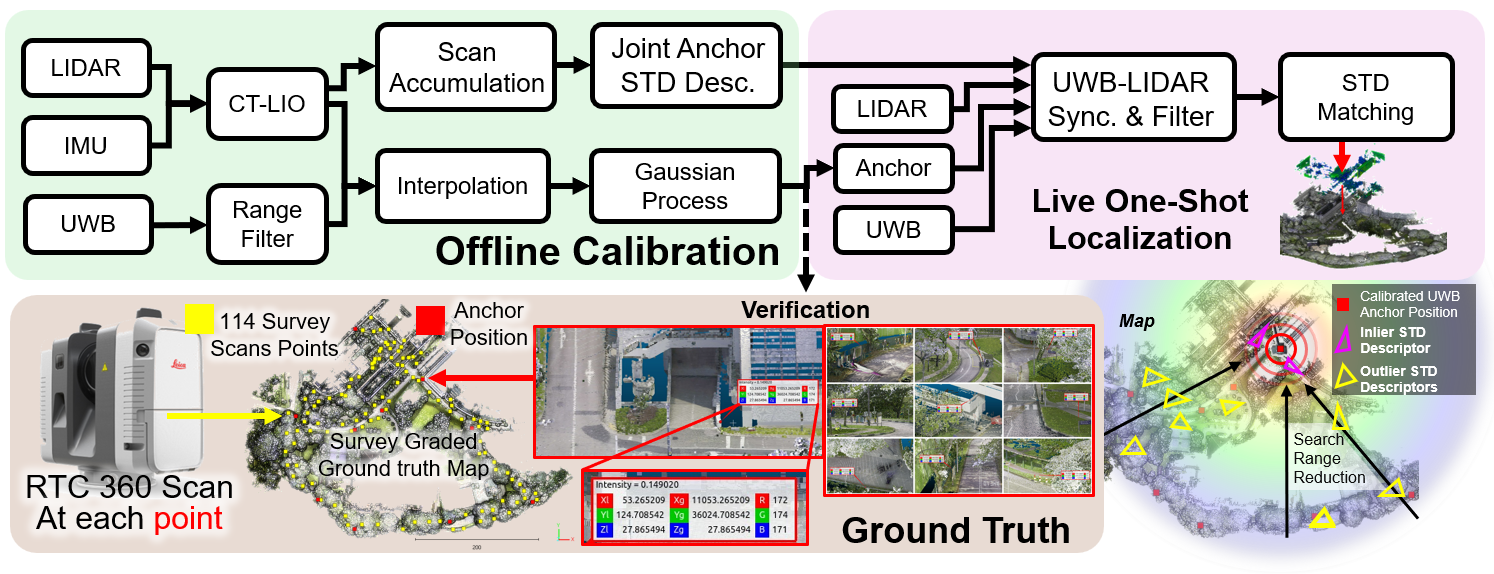}
\vspace{-2.5em}
\caption{\footnotesize The system leverages CT-LIO to generate UWB samples for non-parametric GP fitting, calibrating anchor positions. Using these calibrated positions, a range-based search is integrated with the descriptor method for one-shot localization. Additionally, over 100 scans were collected to create a prior map for ground truth verification. }
\label{fig:system_overview}
\vspace{-1.7em}
\end{figure*}

\vspace{-5pt}
\section{Problem Statement}

Our eventual goal is to estimate the robot's position in large-scale repetitive environments accurately. To do this, we first obtain UWB measurements \( \zeta \in \mathbb{R} \)  and 3D poses \( \mathbf{p}(t) \in \mathbb{R}^3  \) from a LiDAR SLAM framework, using B-spline interpolation. Then, these interpolated poses and samples are integrated into a GP model to calibrate UWB anchor positions, which in turn improves LiDAR localization in repetitive logistic environments.

\vspace{-5pt}
\subsection{Continuous-Time LiDAR-Inertial Odometry}
For the CT-LIO part, we leverage our previous work, SLICT \cite{nguyen2024eigen}, to generate continuous trajectory representations with discrete LiDAR and IMU inputs. To ensure SLICT remains a bounded problem, the system optimizes only a set of control points within a sliding window over the specified interval \( [t_s, t_e] \).
The state of SLICT \( \mathbf{X} \) can be summarized as follows:
 \vspace{-5pt}
\[
\mathbf{X} = \left(T_w, T_{w+1}, \dots, T_{k+N-2}, \beta^I\right),
\]
where \( T_w, \dots, T_{k+N-2} \) are the control points of the local trajectory with N representing the order of the B-spline and  \( [t_s, t_e] \subset [t_w, t_{k+N-2}] \). \( \beta^I  \in \mathbb{R}^6 \) denotes the IMU biases.

The objective function \( f(\mathbf{X}) \) is formulated to minimize the residuals from LiDAR and IMU measurements:
\[
f(\mathbf{X}) = \sum_{k} \|r^L(\mathbf{X})\|^2_{W^L} + \sum_{k} \|r^I(\mathbf{X})\|^2_{W^I},
\]
\vspace{-5pt}

where \( r^L(\mathbf{X}) \) and \( r^I(\mathbf{X}) \) represent the LiDAR and IMU residuals, respectively, and \( W^L \) and \( W^I \) are weight matrices. The cost is minimized using an on-manifold optimization solver that exclusively relies on the Eigen library for the highest efficiency  \cite{nguyen2024eigen}.
\vspace{-5pt}
\subsection{Interpolated UWB poses}
With LiDAR running at 10Hz and UWB running at 20Hz, there are time bias and net frequency differences in the time of acquisition. 
To obtain the exact UWB sample pose \( T(t) = (R(t), \mathbf{p}(t)) \) at any given UWB sampling time \( t \), we perform B-spline interpolation using the control points \( T_i \). \( R(t) \) is the rotation matrix omitted in the calibration process. To interpolate the pose \( T(t) \) at a given time \( t \), the process begins by identifying the knot interval \( [t_i, t_{i+1}) \) such that \( t \) lies within this interval. The normalized time \( s \) within the interval is then computed as \( s = \frac{t - t_i}{t_{i+1} - t_i} \). Next, the blending coefficients \( \lambda_j \) are calculated from the B-spline basis matrix \( B^{(N)} \) and the powers of \( s \) by applying the equation \( [\lambda_0, \lambda_1, \dots, \lambda_{N-1}]^\top = B^{(N)}[1, s, \dots, s^{N-1}]^\top \). These coefficients determine the influence of each control point on the interpolated pose. Finally, the position component \( \mathbf{p}(t) \) of the sample pose can be interpolated as:
\vspace{-5pt}
\[
\mathbf{p}(t) = \mathbf{p}_i + \sum_{j=1}^{N-1} \lambda_j \left(\mathbf{p}_{i+j} - \mathbf{p}_{i+j-1}\right).
\]
This process results in the interpolated pose \( T(t) \) at time \( t \), aligning the robot’s estimated trajectory with specific timestamps.

\subsection{Gaussian Process for UWB Anchor Calibration}

GP \cite{kramer2016scikit} provides a non-parametric framework for determining the underlying distribution of UWB anchors in environment $\xi \subset \mathbb{R}^3$. The unknown distribution is modeled as a continuous function $f(\mathbf{p})$, which represents the range to UWB anchor at a position $\mathbf{p}$, $\mathbf{p} \subset \xi$.  Given a set of samples \( \mathbf{p} \),  generated from SLICT, we define a GP: \(  f(\mathbf{p}) \sim \mathcal{GP}(m(\mathbf{p}), k(\mathbf{p}, \mathbf{p}')) \) that models the relationship between these SLAM-derived poses and the UWB ranging values with set of means and covariance kernel. Here, \( m(\mathbf{p}) \) is the mean function, typically set to zero in the absence of prior information about the anchor positions. \( k(\mathbf{p}, \mathbf{p}')) \) is the covariance function (kernel), defining the relationship between different positions $\mathbf{p}$ and  $\mathbf{p}'$. 

To accurately capture the spatial relationships, we employ a  Matérn 3/2 kernel \cite{matern1960spatial}, chosen for its adaptability in handling varying degrees of smoothness in spatial data:
\[
k(\mathbf{p}, \mathbf{p}') = \frac{2^{1-\nu}}{\Gamma(\nu)} \left(\frac{\sqrt{2\nu} \|\mathbf{p} - \mathbf{p}'\|}{\ell}\right)^\nu K_\nu\left(\frac{\sqrt{2\nu} \|\mathbf{p} - \mathbf{p}'\|}{\ell}\right),
\]
where \( \nu \) and \( \ell \) are the smoothness and length-scale parameters, respectively, which are adjusted based on the environmental complexity and data variability.
The predictive mean \( \mu(\mathbf{p}_*) \) and covariance \( \sigma^2(\mathbf{p}_*) \) of the GP can be obtained as:
\begin{align}
\mu(\mathbf{p}_*) & = k(\mathbf{p}_*, \mathbf{P}) [k(\mathbf{P}, \mathbf{P}) + \sigma^2 I]^{-1} \zeta, \notag\\
\sigma^2(\mathbf{p}_*) & = k(\mathbf{p}_*, \mathbf{p}_*) - k(\mathbf{p}_*, \mathbf{P}) [k(\mathbf{P}, \mathbf{P}) + \sigma^2 I]^{-1} k(\mathbf{P}, \mathbf{p}_*), \notag
\end{align}
where \( \mathbf{p}_* \) represents the pose to be inferred based on known data sampling points \( \mathbf{P} \). 
$\zeta$ is the corresponding UWB signal measurement.

We use an iterative GP to refine predictions by first fitting initial points at a coarse scale, then sampling within a rectangular cuboid region at regular intervals to account for unknown positions. The top 10 sampled values are averaged to update the model. During training, layered random sampling ensures data coverage across the region, improving generalization and convergence towards optimal results. This calibration process enables more accurate anchor positioning, which is critical for improving overall localization performance.

\subsection{One-shot Localization}
In this work, we modify the Stable Triangle Descriptor (STD) \cite{yuan2023std} to prioritize matching scenes close to known UWB anchors. STD alone relies on triangle descriptors, which are prone to errors in repetitive scenes. By integrating calibrated UWB anchors with STD, we aim to mitigate false-positive matches in large, repetitive environments.

UWB measurements are divided into \( \iota_{i,j} \in \mathbb{R}\) discrete range zones, where each zone \( Z_{i,j} \) corresponds to a radius \( r_{i,j} \) from anchor \( a_j \) with an adjusted boundary \( (r_{i-1,j} - \delta) \leq \| p - a_j \| < (r_{i,j} + \delta) \). Using calibration and LiDAR data, we extract \( \iota_{i,j} \) prior descriptors around each anchor \( A_j \). During operation, the current STD descriptor is matched with prior descriptors, determined by the measured UWB distance \( d_j \). The bias \( \delta \in \mathbb{R} \) expands the search region for better STD descriptor representation, improving robustness and reducing errors in repetitive scenes. The parameters for bias and the number of radii are chosen empirically.

\section{Experiment and Result}
\subsection{Real World Dataset}
This work aims to address large-scale environments like Seaports by improving the cost-efficiency of UWB localization. Due to restricted access to operational seaports and the lack of publicly available datasets that include large-scale LiDAR, IMU, and UWB data, we turn to some alternative real-world datasets for validation. 

To validate our approach in real-world environments with the constraint, we conducted experiments using the MCD dataset \cite{nguyen2024mcd}, which covers an area of approximately 600 x 450 m². This real-world dataset mirrors the scale and complexity of a seaport environment, with repetitive scenes that challenge the simple STD descriptor, making it an ideal testbed for our approach.

The original MCD dataset lacked anchor location measurements from other methods. To address this, we collected around 20 minutes of GPS data at each anchor point and averaged the positions to improve accuracy. In some regions, GPS lock required over two hours in heavily occluded areas. In one extreme case, no valid GPS signal was received all day. The resulting available GPS data was aligned with a survey map ground truth using Iterative Closest Point (ICP) for coordinate transformation.  
\subsection{Experiment Setup}
Experiments were conducted on a desktop PC equipped with an i9-13900K CPU and an RTX 4090 GPU for the calibration phase. For the localization experiment, a Gen 11 i7 NUC was used to simulate edge industrial usage scenarios. The ground truth is captured by a map constructed by hundreds of survey scans and is accurate up to cm level. A set of Here+ RTK GPS bases with GPS receivers was deployed to collect precise GPS data, which was later integrated into the existing dataset for comparison. RTK base stations were placed strategically on the rooftop of the NTU S2 building to ensure optimal signal reception and accuracy under the given conditions.  The experiment included 10 Noop Loop LTP-A TOF-type UWB anchors and a tag covering an area of approximately 600 x 450 m². The trajectory data was primarily collected using an all-terrain vehicle (ATV) equipped with two UWB tags mounted on its body. However, only one UWB tag was used in this experiment, as the current GP formulation does not account for the differential angle of arrival between the tags. 

\subsection{Calibration Baseline Methods}
Several open-source implementations are available for the UWB calibration. However, many of these methods are hardware-dependent, such as the Decawave UWB system \cite{sidorenko2020decawave}. From the range of possible solutions, we selected the baseline listed below:\\
\textbf{RTK-GPS }is one of the most intuitive solutions for large-scale UWB calibration. However, many existing solutions \cite{de2019range} rely on RTK-GPS, which requires a clear line of sight to the target GPS. In dense forests or areas surrounded by buildings, this method performs poorly, as the line of sight is frequently obstructed. As a result, the system often defaults to standard GPS mode, making it highly susceptible to occlusion and resulting in reduced accuracy.\\
\textbf{One-Key-Calibration (OKC)} is a UWB calibration method provided by Noop Loop that employed a least squares-based approach for anchor calibration\cite{hol2010ultra}.  The method aims to simplify the calibration process, providing an efficient way to position UWB anchors accurately by minimizing error through least squares optimization.\\
\textbf{Continuous-Time UWB-IMU Fusion (CT-UWB)} is the state-of-the-art method \cite{li2023continuous} that fuses UWB with IMU data to jointly estimate anchor positions using self-excitation with the help of IMU-based on manifold preintegration. It utilizes B-splines to represent the trajectory, optimizing only the quaternion and control points, which significantly increase the accuracy and smoothness. This approach has shown great potential in calibrating anchor positions and accurately estimating self-position, offering a more efficient and scalable solution for environments where traditional methods may struggle. In principle, it is similar to our proposed methods.

\subsection{One-shot Localization Baseline Methods}
Once calibration is achieved, we demonstrate its effectiveness in improving one-shot localization performance. For the \textbf{localization} using UWB calibration prior, we included two configurations that produced successful calibration results alongside the baseline STD descriptor performance. Additionally, we compared the UWB-only localization method provided by Noop Loop with ground truth anchor position manually encoded and the VN-200 GPS-IMU fusion results for a comprehensive evaluation of global one-shot localization. 

The primary aim of this section is not to introduce a new descriptor or loop closure detection method but to show how the proposed anchor calibration enhances global localization methods. UWB in NLoS conditions is often considered ineffective as UWB trilateration typically fails. And STD also failed easily in repetitive environment. However, we demonstrate its potential as a standalone plugin system that can be seamlessly integrated with existing global localization methods, improving both accuracy and robustness, especially in scenarios where other systems struggle to perform reliably or face limitations.


\subsection{Metric Evaluation:} For anchor calibration, the measurements include the average pose error (APE) error of the anchor position and the time taken for capture and processing. For one-shot localization, the measurements are APE, success rate, and average localization time. A match is considered when the error between the predicted and true poses is under 8.5 meters, at which point both the error and success rate are calculated. In the LiDAR-based approach, a match is also confirmed if the angular error is below 10 degrees. If no match is found or fails to localize, the error is not counted for any approach. For GPS and UWB one-shot localization methods, the processing time is not counted.

\begin{table*}
\centering
\small
\caption{ Offline Calibration Result. The best results are highlighted in \textbf{bold}. $\overline{A}$ denotes the average error in m.}
\renewcommand{\arraystretch}{1.5} 
\label{tab:comparison}
\begin{tabular}{ccccccccccccc} 
\toprule
\hline
 &\textbf{A0}$\downarrow$ &\textbf{A1}$\downarrow$   & \textbf{A2}$\downarrow$   & \textbf{A3}$\downarrow$  & \textbf{A4}$\downarrow$   & \textbf{A5}$\downarrow$   & \textbf{A6}$\downarrow$   & \textbf{A7}$\downarrow$  & \textbf{A8}$\downarrow$   & \textbf{A9}$\downarrow$   & $\overline{ \textbf{A} }\downarrow$ & \textbf{Time(min)}$\downarrow$ \\ 
\hline
GPS    & -$^*$  & 42.89 & 37.99 & 39.71 & 43.98 & 38.41 & 38.06 & \textbf{1.01} & 4.78 & 5.77 & 25.26    &  193     \\
OKC \cite{hol2010ultra} & -    & -     & -     & -     & -     & -     & -     & -    & -    & -    & - & -              \\
CT-UWB \cite{li2023continuous} & -  & -     & -     & -     & -     & -     & -     & -    & -    & -    & -  & -             \\
Proposed & - & \textbf{3.77}  & \textbf{0.94}  & \textbf{3.64}  & \textbf{4.91}  & \textbf{1.38}  & \textbf{2.02}  & 1.31 & \textbf{1.32} & \textbf{1.02} &\textbf{ 2.031}   & \textbf{55.64}        \\
\hline
\bottomrule
\end{tabular}
\\
\vspace{3pt}
\footnotesize{$*$If the error exceeds 10\% of map size or the system drifts, we denote the result as '-' to indicate a failure}.
\vspace{-12pt}
\end{table*}

\subsection{Result and Discussion}


\begin{table*}
\centering
\small
\caption{Comparison of localization performances with UWB anchor information. Prior trajectory and scan are selected from NTU\_Night\_08 Sequence, which covers most of the area. The testing sequence overlaps with the prior but with offset.}
\renewcommand{\arraystretch}{1.3} 
\begin{tabular}{lccccccccc} 
\toprule
\hline
         & \multicolumn{2}{c}{NTU\_Night\_04} & \multicolumn{2}{c}{NTU\_Day\_10} & \multicolumn{2}{c}{NTU\_Night\_13} & \multicolumn{3}{c}{Average}  \\
\hline         & Success Rate & APE              & Success Rate & APE               & Success Rate & APE                 & ~Success Rate & APE   & Time       \\

\hline
UWB  & 0\%          & -                & 0\%          & -                 & 0\%          & -                   & 0\%           & - & -           \\
GPS  & \underline{62.34}\%        & 7.398                & 0\%          & -                 & 0\%          & -                   & 20.78\%           & 7.39       & -     \\
STD      & 39.26\%      & 0.617             & 45.01\%      & 0.453              & 36.82\%      & 0.539                & 40.36\%       & 0.536
       & 33.27 ms  \\
STD+$\mathbf{p}_{gps}$  & 53.92\%      & \underline{0.562}             &\underline{52.85}\%      & \underline{0.418}           & \underline{41.99}\%      & \underline{0.482}               & \underline{49.59}\%       & \underline{0.487}      & \textbf{11.28} ms   \\
STD+$\mathbf{p}_{ours}$ & \textbf{72.52}\%      & \textbf{0.371}             & \textbf{75.58}\%      & \textbf{0.358}              & \textbf{68.71}\%      & \textbf{0.453}                & \textbf{72.26}\%       & \textbf{0.394}       & \underline{11.42} ms  \\
\bottomrule
\end{tabular}
\label{tab:localization}
\\
\vspace{5pt}
\footnotesize{ $\mathbf{p}_{gps}$ denotes anchor position obtained from GPS,  $\mathbf{p}_{our}$ denotes anchor position obtained from our proposed solution.}\\
\vspace{-15pt}
\end{table*}

\begin{figure}[t]
    \centering
    \setlength{\abovecaptionskip}{-0.05cm}
    \includegraphics[width=0.48\textwidth]{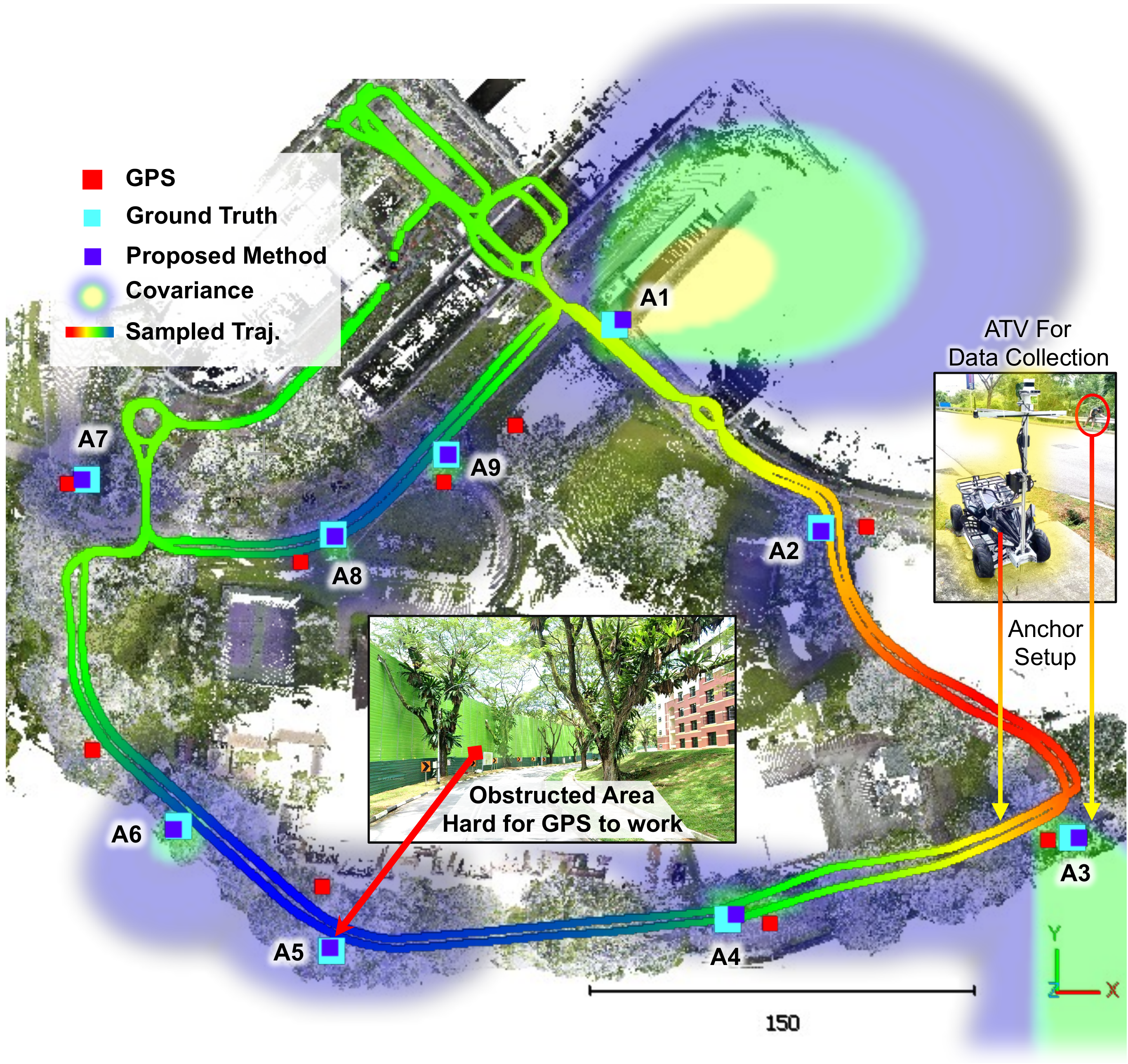}
     \vspace{-5pt}
    \caption{\footnotesize  The proposed solution achieves better prediction accuracy in real-world, large-scale UWB experiments.}
    \label{fig:calibrationresult}
    \vspace{-2em}
\end{figure}

The calibration results are shown in Tab. \ref{tab:comparison} and Fig. \ref{fig:calibrationresult}.  The calibration results indicate that most existing open-source tools and solutions struggle in such large environments. For OKC, issues with larger scales are well-understood, and even within a 50m LoS range, the system often fails to handle the calibration effectively. The CT-UWB-based method initially showed potential, but cumulative IMU drift caused the error to grow uncontrollably, and the error went off the chart. GPS performance was similarly poor, with tall buildings obstructing signal reception and leading to frequent GPS lock failures. Even with an RTK GPS base station, the GPS receiver struggled due to the lack of a clear LoS to satellites. 
During data collection, we placed a GPS unit at the UWB locations and waited hours for a lock, but the readings were inconsistent, with significant drift. For Anchor 0, even after a full day of GPS data collection, there is no GPS-lock which render it impossible to measure with GPS. For the rest of anchor, it took 3 hours to collect GPS data at different locations for the working anchor. \\
\indent In contrast, the proposed method, using SLICT, provided efficient and accurate pose estimates, achieving an average anchor calibration accuracy of around 2 meters. The remaining error can be attributed to odometry drift and UWB biases at longer ranges in tropical humid conditions. The only exception was Anchor 0, which was placed at B5 inside the building, while trajectory samples were collected around B6, leading to the failure of the GP to locate its position. The overall process was efficient, as we spent 5 minutes riding the ATV to each location, followed by about 50 minutes of UWB anchor calibration process.

As shown in Tab. \ref{tab:localization}, We tested the UWB-assisted one-shot localization scheme, leveraging UWB anchor to narrow the search area, which improved matching success rates, reduced processing time, and lowered errors. These results confirm that the proposed method enables efficient one-shot localization with a single NUC-embedded PC. 
While some semantic solutions \cite{yin2023segregator,yin2024outram} can achieve above 90\% success rate, they require a costly GPU platform to handle the semantic prediction. Without a GPU, our approach delivers the best accuracy and efficiency for large-scale UWB-assisted localization. 
For some sequences, GPS localization can achieve a similar localization success rate. However, the GPS error is typically much larger in localization, and in most other cases, GPS fails to function effectively.
We noticed an interesting phenomenon: when GPS remains static for long periods, the sampling errors are not always small. While collecting GPS trajectories using onboard GPS-IMU fusion, we observed that despite frequent failures, there were times when the error was smaller than with static GPS alone. This suggests that IMU can improve accuracy when there are enough accurate GPS samples to perform the fusion effectively.

\vspace{-10pt}
\section{Lessons Learnt}
In this work, Several critical mistakes were encountered that impacted the overall performance. \\
\noindent\textbf{Impractical Anchor Placement:} Anchor 0 was placed on the B5 level of the NTU S1 building, where GPS calibration was not feasible. Additionally, the ATV never moves near anchor 0 (e.g., within 20 meters), as the UGV operated on B6, leading to poor accuracy of the proposed method in its calibration result. \\
\noindent\textbf{UWB Intrinsic Calibration:} 
During extrinsic calibration, we found biases in UWB range measurements when compared to LiDAR range ground truth, causing inaccuracies. This issue, noted in prior literature \cite{de2019range} but overlooked in our data collection, highlights the need to recalibrate UWB Intrinsic parameters for different environments. Without proper calibration, environmental factors like humidity can introduce significant measurement errors, affecting system reliability.

\section{Limitation and future work}

\noindent\textbf{Proven Optimality:} Like the original STD descriptor, which uses a rule-based heuristic method, our approach integrates additional rules, making parameter tuning challenging and mathematically proving optimality difficult. Currently, parameters can be manually tuned or optimized using a reinforcement learning framework \cite{shen2018intelligent} for specific use cases. In the future, we aim to develop a mathematically provable method for the UWB-based solution to ensure guaranteed localization performance.\\
\noindent\textbf{Limited Area Coverage:} We tested our approach in a sizable area, though smaller than the world's largest ports, which span several kilometers. Scaling was limited by restricted access and the need for UWB anchor installations. LiDAR odometry drift also requires correction mechanisms. Future work will involve real-world testing using large-scale bundle adjustment \cite{li2024pss} and divide-and-conquer methods for regional anchor localization \cite{barajas2018multivariate}, similar to calibrating and merging multiple MCD datasets. \\
\noindent\textbf{Reliance on High-Performance LiDAR:} Another issue is the reliance on the costly 128-line LiDAR, which limits its feasibility for large-scale operations. Most service autonomous mobile robots (AMR) use 16-line LiDARs or vision-based approaches. Future work will focus on improving accuracy and cost efficiency by estimating dynamic biases and replacing CT-LIO with a more affordable vision-based approach, such as AirVO \cite{xu2023airvo}, \cite{chen2024salient}, \cite{airSLAM}, to reduce reliance on expensive LiDARs.

\section{Conclusion}
We propose a UWB calibration method to address NLoS issues using SLICT to generate a continuous-time trajectory. By sampling UWB acquisition poses and applying an iterative GP, we successfully calibrated anchors in large 600x450 m² environments with approximately 2-meter accuracy. Even in NLoS conditions, where traditional UWB localization fails, the calibrated anchor positions serve as a standalone plugin to improve one-shot localization success rates in repetitive and large-scale environments. We demonstrate that container-carrying AMRs can achieve accurate one-shot localization in repetitive environments using sparsely populated UWBs and LiDAR onboard at minimal cost.

\bibliographystyle{IEEEtran}
\bibliography{mybib}


\end{document}